# Handling Sparse Data by Successive Abstraction

Christer Samuelsson
Universität des Saarlandes, FR 8.7, Computerlinguistik
Postfach 1150, D-66041 Saarbrücken, Germany
Internet: christer@coli.uni-sb.de

## Abstract

A general, practical method for handling sparse data that avoids held-out data and iterative reestimation is derived from first principles. It has been tested on a part-of-speech tagging task and outperformed (deleted) interpolation with context-independent weights, even when the latter used a globally optimal parameter setting determined a posteriori.

## 1 Introduction

Sparse data is a perennial problem when applying statistical techniques to natural language processing. The fundamental problem is that there is often not enough data to estimate the required statistical parameters, i.e., the probabilities, directly from the relative frequencies. This problem is accentuated by the fact that in the search for more accurate probabilistic language models, more and more contextual information is added, resulting in more and more complex conditionings of the corresponding conditional probabilities. This in turn means that the number of observations tends to be quite small for such contexts. Over the years, a number of techniques have been proposed to handle this problem.

One of two different main ideas behind these techniques is that complex contexts can be generalized, and data from more general contexts can be used to improve the probability estimates for more specific contexts. This idea is usually referred to as back-off smoothing, see (Katz 1987). These techniques typically require that a separate portion of the training data be held out from the parameter-estimation phase and saved for determining appropriate back-off weights. Furthermore, determining the back-off weights usually requires resorting to a time-consuming iterative reestimation procedure. A typical example of such a technique is "deleted interpolation", which is described in Section 5.1 below.

The other main idea is concerned with improving the estimates of low-frequency, or no-frequency, outcomes apparently without trying to generalize the conditionings. Instead, these techniques are based on considerations of how population frequencies in general tend to behave. Examples of this are expected likelihood estimation (ELE), see Section 5.2 below, and Good-Turing estimation, see (Good 1953).

We will here derive from first principles a practical method for handling sparse data that does not need separate training data for determining the back-off weights and which lends itself to direct calculation, thus avoiding time-consuming reestimation procedures.

## 2 Linear Successive Abstraction

Assume that we want to estimate the conditional probability $P(x \mid C)$ of the outcome $x$ given a context $C$ from the number of times $N_x$ it occurs in $N = |C|$ trials, but that this data is sparse. Assume further that there is abundant data in a more general context $C' \supset C$ that we want to use to get a better estimate of $P(x \mid C)$. The idea is to let the probability estimate $\tilde{P}(x \mid C)$ in context $C$ be a function $g$ of the relative frequency $f(x \mid C)$ of the outcome $x$ in context $C$ and the probability estimate $\tilde{P}(x \mid C')$ in context $C'$:

$$\tilde{P}(x \mid C) = g(f(x \mid C), \tilde{P}(x \mid C'))$$

Let us generalize this scenario slightly to the situation were we have a sequence of increasingly more general contexts $C_m \subset C_{m-1} \subset \ldots \subset C_1$, i.e., where there is a linear order of the various contexts $C_k$. We can then build the estimate of $P(x \mid C_k)$ on the relative frequency $f(x \mid C_k)$ in context $C_k$ and the previously established estimate of $P(x \mid C_{k-1})$. We call this method *linear successive abstraction*. A simple example is estimating the probability $P(x \mid l_{n-j+1}, \ldots, l_n)$ of word class $x$ given $l_{n-j+1}, \ldots, l_n$, the last $j$ letters of a word $l_1, \ldots, l_n$. In this case, the estimate will be based on the relative frequencies $f(x \mid l_{n-j+1}, \ldots, l_n), \ldots, f(x \mid l_n), f(x)$.

We will here consider the special case when the function $g$ is a weighted sum of the relative frequency and the previous estimate, appropriately

renormalized:

$$\tilde{P}(x \mid C_k) = \frac{f(x \mid C_k) + \theta \tilde{P}(x \mid C_{k-1})}{1 + \theta}$$

We want the weight $\theta$ to depend on the context $C_k$, and in particular be proportional to some measure of how spread out the relative frequencies of the various outcomes in context $C_k$ are from the statistical mean. The variance is the quadratic moment w.r.t. the mean, and is thus such a measure. However, we want the weight to have the same dimension as the statistical mean, and the dimension of the variance is obviously the square of the dimension of the mean. The square root of the variance, which is the standard deviation, should thus be a suitable quantity. For this reason we will use the standard deviation in $C_k$ as a weight, i.e., $\theta = \sigma(C_k)$. One could of course multiply this quantity with any reasonable real constant, but we will arbitrarily set this constant to one, i.e., use $\sigma(C_k)$ itself.

In linguistic applications, the outcomes are usually not real numbers, but pieces of linguistic structure such as words, part-of-speech tags, grammar rules, bits of semantic tissue, etc. This means that it is not quite obvious what the standard deviation, or the statistical mean for that matter, actually should be. To put it a bit more abstractly, we need to calculate the standard deviation of a non-numerical random variable.

### 2.1 Deriving the Standard Deviation

So how do we find the standard deviation of a non-numerical random variable? One way is to construct an equivalent numerical random variable and use the standard deviation of the latter. This can be done in several different ways. The one we will use is to construct a numerical random variable with a uniform distribution that has the same entropy as the non-numerical one. Whether we use a discrete or continuous random variable is, as we shall see, of no importance.

We will first factor out the dependence on the context size. Quite in general, if $\bar{\xi}_N$ is the sample mean of $N$ independent observations of any numerical random variable $\xi$ with variance $\sigma_0^2$, i.e., $\bar{\xi}_N = \frac{1}{N} \sum_{i=1}^{N} \xi_i$, then

$$\sigma^2 = \mathrm{Var}[\bar{\xi}_N] =$$
$$= \mathrm{Var}[\frac{1}{N}\sum_{i=1}^{N}\xi_i] = \frac{1}{N^2}\sum_{i=1}^{N}\mathrm{Var}[\xi_i] = \frac{\sigma_0^2}{N}$$

In our case, the number of observations $N$ is simply the size of the context $C_k$, by which we mean the number of times $C_k$ occurred in the training data, i.e., the frequency count of $C_k$, which we will denote $|C_k|$. Since the standard deviation is the square root of the variance, we have

$$\sigma(C_k) = \frac{\sigma_0(C_k)}{\sqrt{|C_k|}}$$

Here $\sigma_0$ does not depend on the number of observations in context $C_k$, only on the underlying probability distribution conditional on context $C_k$.

To estimate $\sigma_0(C_k)$, we assume that we have either a discrete uniform distribution on $\{1,\ldots,M\}$ or a continuous uniform distribution on $[0, M]$ that is as hard to predict as the one in $C_k$ in the sense that the entropy is the same. The entropy $\mathrm{H}[\xi]$ of a random variable $\xi$ is the expectation value of the logarithm of $P(\xi)$. In the discrete case we thus have

$$\mathrm{H}[\xi] = \mathrm{E}[-\ln P(\xi)] = \sum_i -P(x_i) \ln P(x_i)$$

Here $P(x_i)$ is the probability of the random variable $\xi$ taking the value $x_i$, which is $\frac{1}{M}$ for all possible outcomes $x_i$ and zero otherwise. Thus, the entropy is $\ln M$:

$$\sum_i -P(x_i) \ln P(x_i) = \sum_{i=1}^{M} -\frac{1}{M} \ln \frac{1}{M} = \ln M$$

The continuous case is similar. We thus have that

$$\ln M = \mathrm{H}[C_k] \quad \text{or} \quad M = e^{\mathrm{H}[C_k]}$$

The variance of these uniform distributions is $\frac{M^2}{12}$ in the continuous case and $\frac{M^2-1}{12}$ in the discrete case. We thus have

$$\sigma_0(C_k) = \frac{M}{\sqrt{12}} = \frac{1}{\sqrt{12}\,M^{-1}} = \frac{1}{\sqrt{12}\,e^{-\mathrm{H}[C_k]}}$$

Unfortunately, the entropy $\mathrm{H}[C_k]$ depends on the probability distribution of context $C_k$ and thus on $\sigma_0(C_k)$. Since we want to avoid trying to solve highly nonlinear equations, and since we have access to an estimate of the probability distribution of context $C_{k-1}$, we will make the following approximation:

$$\sigma(C_k) \approx \frac{\sigma_0(C_{k-1})}{\sqrt{|C_k|}} = \frac{1}{\sqrt{|C_k|}\sqrt{12}\,e^{-\mathrm{H}[C_{k-1}]}}$$

It is starting to look sensible to specify $\sigma^{-1}$ instead of $\sigma$, i.e., instead of $\frac{f + \sigma \tilde{P}}{1 + \sigma}$, we will write $\frac{\sigma^{-1} f + \tilde{P}}{\sigma^{-1} + 1}$.

### 2.2 The Final Recurrence Formula

We have thus established a recurrence formula for the estimate of the probability distribution in context $C_k$ given the estimate of the probability distribution in context $C_{k-1}$ and the relative frequencies in context $C_k$:

$$\tilde{P}(x \mid C_k) = \qquad (1)$$
$$= \frac{\sigma(C_k)^{-1} f(x \mid C_k) + \tilde{P}(x \mid C_{k-1})}{\sigma(C_k)^{-1} + 1}$$

and

$$\sigma(C_k)^{-1} = \sqrt{12}\,\sqrt{|C_k|}\,e^{-\mathrm{H}[C_{k-1}]}$$

We will start by estimating the probability distribution in the most general context $C_1$, if necessary directly from the relative frequencies. Since this is the most general context, this will be the context with the most training data. Thus it stands the best chances of the relative frequencies being acceptably accurate estimates. This will allow us to calculate an estimate of the probability distribution in context $C_2$, which in turn will allow us to calculate an estimate of the probability distribution in context $C_3$, etc. We can thus calculate estimates of the probability distributions in all contexts $C_1, \ldots, C_m$.

We will next consider some examples from part-of-speech tagging.

## 3 Examples from PoS Tagging

Part-of-speech (PoS) tagging consists in assigning to each word of an input text a (set of) tag(s) from a finite set of possible tags, a tag palette or a tag set. The reason that this is a research issue is that a word can in general be assigned different tags depending on context. In statistical tagging, the relevant information is extracted from a training text and fitted into a statistical language model, which is then used to assign the most likely tag to each word in the input text.

The statistical language model usually consists of lexical probabilities, which determine the probability of a particular tag conditional on the particular word, and contextual probabilities, which determine the probability of a particular tag conditional on the surrounding tags. The latter conditioning is usually on the tags of the neighbouring words, and very often on the $N-1$ previous tags, so-called (tag) $N$-gram statistics. These probabilities can be estimated either from a pretagged training corpus or from untagged text, a lexicon and an initial bias. We will here consider the former case.

Statistical taggers usually work as follows: First, each word in the input word string $W_1, \ldots, W_n$ is assigned all possible tags according to the lexicon, thereby creating a lattice. A dynamic programming technique is then used to find tag the sequence $T_1, \ldots, T_n$ that maximizes

$$P(T_1, \ldots, T_n \mid W_1, \ldots, W_n) =$$
$$= \prod_{k=1}^{n} P(T_k \mid T_1, \ldots, T_{k-1}; W_1, \ldots, W_n) \approx$$
$$\approx \prod_{k=1}^{n} P(T_k \mid T_{k-N+1}, \ldots, T_{k-1}; W_k) \approx$$
$$\approx \prod_{k=1}^{n} \frac{P(T_k \mid T_{k-N+1}, \ldots, T_{k-1}) \cdot P(T_k \mid W_k)}{P(T_k)}$$
$$= \prod_{k=1}^{n} \frac{P(T_k \mid T_{k-N+1}, \ldots, T_{k-1}) \cdot P(W_k \mid T_k)}{P(W_k)}$$

Since the maximum does not depend on the factors $P(W_k)$, these can be omitted, yielding the standard statistical PoS tagging task:

$$\max_{T_1, \ldots, T_n} \prod_{k=1}^{n} P(T_k \mid T_{k-N+1}, \ldots, T_{k-1}) \cdot P(W_k \mid T_k)$$

This is well-described in for example (DeRose 1988).

We thus have to estimate the two following sets of probabilities:

- **Lexical probabilities:**
  The probability of each tag $T^i$ conditional on the word $W$ that is to be tagged, $P(T^i \mid W)$. Often the converse probabilities $P(W \mid T^i)$ are given instead, but we will for reasons soon to become apparent use the former formulation.

- **Tag $N$-grams:**
  The probability of tag $T^i$ at position $k$ in the input string, denoted $T_k^i$, given that tags $T_{k-N+1}, \ldots, T_{k-1}$ have been assigned to the previous $N-1$ words. Often $N$ is set to two or three, and thus bigrams or trigrams are employed. When using trigram statistics, this quantity is $P(T_k^i \mid T_{k-2}, T_{k-1})$.

### 3.1 $N$-gram Back-off Smoothing

We will first consider estimating the $N$-gram probabilities $P(T_k^i \mid T_{k-N+1}, \ldots, T_{k-1})$. Here, there is an obvious sequence of generalizations of the context $T_{k-N+1}, \ldots, T_{k-1}$ with a linear order, namely $T_{k-N+1}, \ldots, T_{k-1} \subset T_{k-N+2}, \ldots, T_{k-1} \subset \ldots \subset T_{k-1} \subset \Omega$, where $\Omega$ means "no information", corresponding to the unigram probabilities. Thus we will repeatedly strip off the tag furthest from the current word and use the estimate of the probability distribution in this generalized context to improve the estimate in the current context. This means that when estimating the $(j+1)$-gram probabilities, we back off to the estimate of the $j$-gram probabilities.

So when estimating $P(T_k^i \mid T_{k-j}, \ldots, T_{k-1})$, we simply strip off the tag $T_{k-j}$ and apply Eq. (1):

$$\tilde{P}(T_k^i \mid T_{k-j}, \ldots, T_{k-1}) =$$
$$= \frac{\sigma^{-1} f(T_k^i \mid T_{k-j}, \ldots, T_{k-1})}{\sigma^{-1} + 1} +$$
$$+ \frac{\tilde{P}(T_k^i \mid T_{k-j+1}, \ldots, T_{k-1})}{\sigma^{-1} + 1}$$

and

$$\sigma^{-1} = \sqrt{12} \sqrt{|T_{k-j}, \ldots, T_{k-1}|} \, e^{-\mathrm{H}[T_{k-j+1}, \ldots, T_{k-1}]}$$

### 3.2 Handling Unknown Words

We will next consider improving the probability estimates for unknown words, i.e., words that do not occur in the training corpus, and for which we therefore have no lexical probabilities. The same technique could actually be used for improving the estimates of the lexical probabilities of words that do occur in the training corpus. The basic idea is that there is a substantial amount of information in the word suffixes, especially for languages with a richer morphological structure than English. For this reason, we will estimate the probability distribution conditional on an unknown word from the statistical data available for words that end with the same sequence of letters. Assume that the word consists of the letters $l_1, \ldots, l_n$. We want to know the probabilities $P(T^i \mid l_1, \ldots, l_n)$ for the various tags $T^i$.[1] Since the word is unknown, this data is not available. However, if we look at the sequence of generalizations of "ending with same last $j$ letters", here denoted $l_{n-j+1}, \ldots, l_n$, we realize that sooner or later, there will be observations available, in the worst case looking at the last zero letters, i.e., at the unigram probabilities.

So when estimating $P(T^i \mid l_{n-j+1}, \ldots, l_n)$, we simply omit the $j$th last letter $l_{n-j+1}$ and apply Eq. (1):

$$\tilde{P}(T^i \mid l_{n-j+1}, \ldots, l_n) =$$
$$= \frac{\sigma^{-1} f(T^i \mid l_{n-j+1}, \ldots, l_n)}{\sigma^{-1} + 1} +$$
$$+ \frac{\tilde{P}(T^i \mid l_{n-j+2}, \ldots, l_n)}{\sigma^{-1} + 1}$$

and

$$\sigma^{-1} = \sqrt{12} \sqrt{|l_{n-j+1}, \ldots, l_n|} e^{-H[l_{n-j+2}, \ldots, l_n]}$$

This data can be collected from the words in the training corpus with frequencies below some threshold, e.g., words that occur less than say ten times, and can be indexed in a tree on reversed suffixes for quick access.

## 4 Partial Successive Abstraction

If there is only a partial order of the various generalizations, the scheme is still viable. For example, consider generalizing symmetric trigram statistics, i.e., statistics of the form $P(T \mid T_l, T_r)$. Here, both $T_l$, the tag of the word to the left, and $T_r$, the tag of the word to the right, are one-step generalizations of the context $T_l, T_r$, and both have in turn the common generalization $\Omega$ ("no information"). We modify Eq. (1) accordingly:

$$\tilde{P}(T \mid T_l, T_r) = \frac{\sigma(T_l, T_r)^{-1} f(T \mid T_l, T_r)}{\sigma(T_l, T_r)^{-1} + 1} +$$
$$+ \frac{1}{2} \frac{\tilde{P}(T \mid T_l) + \tilde{P}(T \mid T_r)}{\sigma(T_l, T_r)^{-1} + 1}$$

and

$$\tilde{P}(T \mid T_l) = \frac{\sigma(T_l)^{-1} f(T \mid T_l) + \tilde{P}(T)}{\sigma(T_l)^{-1} + 1}$$
$$\tilde{P}(T \mid T_r) = \frac{\sigma(T_r)^{-1} f(T \mid T_r) + \tilde{P}(T)}{\sigma(T_r)^{-1} + 1}$$

---
[1] Or really, $P(T^i \mid l_0, l_1, \ldots, l_n)$ where $l_0$ is a special symbol indicating the beginning of the word.

We call this *partial successive abstraction*. Since we really want to estimate $\sigma$ in the more specific context, and since the standard deviation (with the dependence on context size factored out) will most likely not increase when we specialize the context, we will use:

$$\sigma(T_l, T_r) = \frac{1}{\sqrt{|T_l, T_r|}} \min(\sigma_0(T_l), \sigma_0(T_r))$$

In the general case, where we have $M$ one-step generalizations $C'_i$ of $C$, we arrive at the equation

$$\tilde{P}(x \mid C) =$$
$$= \frac{\sigma(C)^{-1} f(x \mid C) + \frac{1}{M} \sum_{i=1}^{M} \tilde{P}(x \mid C'_i)}{\sigma(C)^{-1} + 1}$$

and

$$\sigma(C) = \frac{1}{\sqrt{|C|}} \min_{i \in \{1, \ldots, M\}} \sigma_0(C'_i)$$
$$\sigma_0(C'_i)^{-1} = \sqrt{12} e^{-H[C'_i]}$$

By calculating the estimates of the probability distributions in such an order that whenever estimating the probability distribution in some particular context, the probability distributions in all more general contexts have already been estimated, we can guarantee that all quantities necessary for the calculations are available.

## 5 Relationship to Other Methods

We will next compare the proposed method to, in turn, deleted interpolation, expected likelihood estimation and Katz's back-off scheme.

### 5.1 Deleted Interpolation

Interpolation requires that the training corpus is divided into one part used to estimate the relative frequencies, and a separate held-back part used to cope with sparse data through back-off smoothing. For example, tag trigram probabilities can be estimated as follows:

$$P(T_k^i \mid T_{k-2}, T_{k-1}) \approx \lambda_1 f(T_k^i) +$$
$$+ \lambda_2 f(T_k^i \mid T_{k-1}) + \lambda_3 f(T_k^i \mid T_{k-2}, T_{k-1})$$

Since the probability estimate is a linear combination of the various observed relative frequencies, this is called linear interpolation. The weights $\lambda_j$ may depend on the conditionings, but are required to be nonnegative and to sum to one over $j$. An enhancement is to partition the training set into $n$ parts and in turn perform linear interpolation with each of the $n$ parts held out to determine the back-off weights and use the remaining $n - 1$ parts for parameter estimation. The various back-off weights are combined in the process. This is usually referred to as *deleted interpolation*.

The weights $\lambda_j$ are determined by maximizing the probability of the held-out part of the training data, see (Jelinek & Mercer 1980). A locally optimal weight setting can be found using Baum-Welch reestimation, see (Baum 1972). Baum-Welch reestimation is however prohibitively time-consuming for complex contexts if the weights are allowed to depend on the contexts, while successive abstraction is clearly tractable; the latter effectively determines these weights directly from the same data as the relative frequencies.

## 5.2 Expected Likelihood Estimation

Expected likelihood estimation (ELE) consists in assigning an extra half a count to all outcomes. Thus, an outcome that didn't occur in the training data receives half a count, an outcome that occurred once receives three half counts. This is equivalent to assigning a count of one to the occurring, and one third to the non-occurring outcomes. To give an indication of how successive abstraction is related to ELE, consider the following special case: If we indeed have a uniform distribution with $M$ outcomes of probability $\frac{1}{M}$ in context $C_{k-1}$ and there is but one observation of one single outcome in context $C_k$, then Eq. (1) will assign to this outcome the probability $\frac{\sqrt{12}+1}{\sqrt{12}+M}$ and to the other, non-occurring, outcomes $\frac{1}{\sqrt{12}+M}$. So if we had used 2 instead of $\sqrt{12}$ in Eq. (1), this would have been equivalent to assigning a count of one to the outcome that occurred, and a count of one third to the ones that didn't. As it is, the latter outcomes are assigned a count of $\frac{1}{\sqrt{12}+1}$.

## 5.3 Katz's Back-Off Scheme

The proposed method is identical to Katz's back-off method (Katz 1987) up to the point of suggesting a, in the general case non-linear, retreat to more general contexts:

$$\tilde{P}(x \mid C) = g(f(x \mid C), \tilde{P}(x \mid C'))$$

Blending the involved distributions $f(x \mid C)$ and $\tilde{P}(x \mid C')$, rather than only backing off to $C'$ if $f(x \mid C)$ is zero, and in particular, instantiating the function $g(f, \tilde{P})$ to a weighted sum, distinguishes the two approaches.

## 6 Experiments

A standard statistical trigram tagger has been implemented that uses linear successive abstraction for smoothing the trigram and bigram probabilities, as described in Section 3.1, and that handles unknown words using a reversed suffix tree, as described in Section 3.2, again using linear successive abstraction to improve the probability estimates. This tagger was tested on the Susanne Corpus, (Sampson 1995), using a reduced tag set of 62 tags. The size of the training corpus $A$ was almost 130,000 words. There were three separate test corpora $B$, $C$ and $D$ consisting of approximately 10,000 words each.

| Test corpus | $B$ | | |
|---|---|---|---|
| Tagger | bigram | trigram | HMM |
| Error rate (%) | 4.41 | 4.36 | 4.49 |
| – tag omissions | | 0.67 | |
| – unknown words | 1.36 | 1.20 | 1.52 |
| Unknown words | | 6.18 | |
| Error rate (%) | 22.1 | 19.4 | 24.5 |
| Test corpus | $C$ | | |
| Tagger | bigram | trigram | HMM |
| Error rate (%) | 4.26 | 3.93 | 4.03 |
| – tag omissions | | 0.68 | |
| – unknown words | 1.43 | 1.30 | 1.34 |
| Unknown words | | 7.78 | |
| Error rate (%) | 18.3 | 16.8 | 17.3 |
| Test corpus | $D$ | | |
| Tagger | bigram | trigram | HMM |
| Error rate (%) | 5.14 | 4.81 | 5.13 |
| – tag omissions | | 0.94 | |
| – unknown words | 1.80 | 1.63 | 2.02 |
| Unknown words | | 8.06 | |
| Error rate (%) | 22.3 | 20.2 | 25.0 |

Figure 1: Results on the Susanne Corpus

The performance of the tagger was compared with that of an HMM-based trigram tagger that uses linear interpolation for $N$-gram smoothing, but where the back-off weights do not depend on the conditionings. An optimal weight setting was determined for each test corpus individually, and used in the experiments. Incidentally, this setting varied considerably from corpus to corpus. Thus, this represented the best possible setting of back-off weights obtainable by linear interpolation, and in particular by linear deleted interpolation, when these are not allowed to depend on the context.

In contrast, the successive abstraction scheme determined the back-off weights automatically from the training corpus alone, and the same weight setting was used for all test corpora, yielding results that were at least on par with those obtained using linear interpolation with a globally optimal setting of context-independent back-off weights determined a posteriori. Both taggers handled unknown words by inspecting the suffixes, but the HMM-based tagger did not smooth the probability distributions.

The experimental results are shown in Figure 1. Note that the absolute performance of the trigram tagger is around 96 % accuracy in two cases and distinctly above 95 % accuracy in all cases, which is clearly state-of-the-art results. Since each test corpus consisted of about 10,000 words, and the error rates are between 4 and 5 %, the 5 percent significance level for differences in error rate is between 0.39 and 0.43 % depending on the error rate, and the 10 percent significance level is between 0.32 and 0.36 %.

We see that the trigram tagger is better than the bigram tagger in all three cases and significantly better at significance level 10 percent, but not at 5 percent, in case $C$. So at this significance level, we can conclude that smoothed trigram statistics improve on bigram statistics alone. The trigram tagger performed better than the HMM-based one in all three cases, but not significantly better at any significance level below 10 percent. This indicates that the successive abstraction scheme yields back-off weights that are at least as good as the best ones obtainable through linear deleted interpolation with context-independent back-off weights.

## 7 Summary and Further Directions

In this paper, we derived a general, practical method for handling sparse data from first principles that avoids held-out data and iterative reestimation. It was tested on a part-of-speech tagging task and outperformed linear interpolation with context-independent weights, even when the latter used a globally optimal parameter setting determined a posteriori.

Informal experiments indicate that it is possible to achieve slightly better performance by replacing the expression for $\sigma_0^{-1}(C_k)$ with a fixed global constant (while retaining the factor $\frac{1}{\sqrt{|C_k|}}$, which is most likely a quite accurate model of the dependence on context size). However, the optimal value for this parameter varied more than an order of magnitude, and the improvements in performance were not very large. Furthermore, suboptimal choices of this parameter tended to degrade performance, rather than improve it. This indicates that the proposed formula is doing a pretty good job of approximating an optimal parameter choice. It would nonetheless be interesting to see if the formula could be improved on, especially seeing that it was theoretically derived, and then directly applied to the tagging task, immediately yielding the quoted results.

## Acknowledgements

The work presented in this article was funded by the N3 "Bidirektionale Linguistische Deduktion (BiLD)" project in the Sonderforschungsbereich 314 *Künstliche Intelligenz — Wissensbasierte Systeme*.

I wish to thank greatly Thorsten Brants, Slava Katz, Khalil Sima'an, the audiences of seminars at the University of Pennsylvania and the University of Sussex, in particular Mark Liberman, and the anonymous reviewers of Coling and ACL for pointing out inaccuracies and supplying useful comments and suggestions to improvements.